%% file: main.tex
\documentclass{isprs} 
\usepackage{subfig}
\usepackage{setspace}
\usepackage{geometry} 
\usepackage{epstopdf}
\usepackage[british]{babel} 
\usepackage{multirow}
\usepackage{xcolor}
\usepackage{svg}
\usepackage{url}
\usepackage{hhline}

\begin{document}

\title{Towards Open-Set Semantic Segmentation of Aerial Images}

\author{
Caio C. V. da Silva\textsuperscript{1}, Keiller Nogueira\textsuperscript{1,2}, Hugo N. Oliveira\textsuperscript{1}, Jefersson A. dos Santos\textsuperscript{1}
}

\address{
\textsuperscript{1} Department of Computer Science, Universidade Federal de Minas Gerais,Belo Horizonte, Brazil\\
\{caiosilva, keiller.nogueira, oliveirahugo, jefersson\}@dcc.ufmg.br \\
\textsuperscript{2} Computing Science and Mathematics, University of Stirling, Stirling, FK9 4LA, Scotland, UK \\ kno@cs.stir.uk
}


\newcommand{\todo}[1]{\textcolor{red}{[TODO: #1]}}

\icwg{}
\abstract{
Classical and more recently deep computer vision methods are optimized for visible spectrum images, commonly encoded in grayscale or RGB colorspaces acquired from smartphones or cameras.
A more uncommon source of images exploited in the remote sensing field are satellite and aerial images. However the development of pattern recognition approaches for these data is relatively recent, mainly due to the limited availability of this type of images, as until recently they were used exclusively for military purposes. Access to aerial imagery, including spectral information, has been increasing mainly due to the low cost of drones, cheapening of imaging satellite launch costs, and novel public datasets. Usually remote sensing applications employ computer vision techniques strictly modeled for classification tasks in closed set scenarios. However, real-world tasks rarely fit into closed set contexts, frequently presenting previously unknown classes, characterizing them as open set scenarios. 
Focusing on this problem, this is the first paper to study and develop semantic segmentation techniques for open set scenarios applied to remote sensing images. The main contributions of this paper are: 1) a discussion of related works in open set semantic segmentation, showing evidence that these techniques can be adapted for open set remote sensing tasks; 2) the development and evaluation of a novel approach for open set semantic segmentation. Our method yielded competitive results when compared to closed set methods for the same dataset.
}


\maketitle

\newcommand{\currprop}{0.95\columnwidth} 

\input{sec1_introduction}
\input{sec2_related_works}

\input{sec3_methodology}
\input{sec4_experiments}

\input{sec5_results}

\input{sec6_conclusion}

 \section*{ACKNOWLEDGEMENTS}
 \label{ACKNOWLEDGEMENTS}

 Authors would like to thank NVIDIA for the donation of the GPUs that allowed the execution of our experiments. We also thank CAPES, CNPq, and FAPEMIG for the financial support provided for this research.

{
	\begin{spacing}{1.17}
		\normalsize
		\bibliography{ISPRSguidelines_authors} 
	\end{spacing}
}




\end{document}

%% file: sec1_introduction.tex
\section{Introduction}
\label{sec:introduction}


The main approaches developed in computer vision and digital image processing are focused on data obtained through smartphones, compact cameras, smartwatches, glasses, and so on. Those cameras normally are used to capture images composed of RGB channels, within the visible spectrum.
Aside from this, another source of data exploited by the computer vision domain is satellite or aerial images, which may be comprised of beyond the visible spectrum channels. However, satellite imagery has not been widely used mainly due to its limited availability given that, until recently, they were of exclusive military use.
However, access to aerial images, including spectral information, has been increasing mainly due to the low cost of drones, new civilian satellites, and data sets on various public platforms, leading to an increase on the amount of remote sensing research.



Within the area of remote sensing, most applications work with closed set scenario-based computer vision techniques, wherein models are conceived to learn and predict the same set of classes, ignoring new unknown labels that may arise.
However, the world is not purely closed set, since many scenarios present objects that were not previously known by the model.
These other scenarios would be better explored using open set algorithms \cite{scheirer2013toward}, which are capable of distinguishing new unknown classes from the ones used during training.
This may be especially appropriate for the remote sensing domain given the nature of the images, i.e., many of the plantations, cities or objects that appear in these images are restricted to the place where it was registered.
Training a model capable of categorizing all possible classes and be used in different images is extremely difficult.
Besides that, such scenario cannot be as well controlled as others in traditional computer vision applications, given that there is an undefined amount of objects not known by the model that can be registered on the images.

Open set classification can be described as a task wherein an image can be labeled as belonging to one of the classes learned by the algorithm or as an unknown class if it belongs to any class not learned. The main challenges of open set classification are: (1) high diversity of patterns in the unknown label, since it can aggregate multiple classes that were not present during training; (2) high similarity between known and unknown classes.
For example, an algorithm trained to classify trees on aerial images, may have difficulty in labeling grass as unknown in those images.
Even with a variety of possible uses for open set scenario algorithms, this area is not so explored, mainly when compared to the enormous number of closed set methods. Analyzing the small world of open set techniques, one can observe that most of them perform scene classification, evidencing a considerable gap in semantic segmentation. 

Semantic Segmentation is a task that aims to classify not only the whole image but every pixel in an image accordingly to the classes learned by the algorithm. This is a very complex task given it requires: (1) densely labeled datasets, where each image needs to have all the pixels annotated; 
(2) powerful algorithms, that need to take in account the classification of every single pixel in its decision-making process.
Therefore, open set semantic segmentation can be described as a set of techniques that receive an image as input and outputs a prediction for all pixels, either labeling them within a known class or as belonging to an unknown class. In this work, we introduce the concept of open set semantic segmentation in remote sensing and propose two deep learning methods based on it. Furthermore, the proposed approaches were extensively evaluated using a well-known remote sensing dataset.


In practice, the main contributions of this work are: 
(1) first evidences demonstrating the feasibility of combining semantic segmentation and open set concepts; 
(2) a new method, called OpenPixel, for open set semantic segmentation based on the Pixelwise network \cite{nogueira2016learning}; and 
(3) an adaptation of the proposed OpenPixel method, applying a morphological filter.

The rest of this paper are organized as follows. Section~\ref{sec:related_work} presents the related works, explaining some others techniques existent on the open set scenario and what are the differences to the methods proposed in this paper. Section~\ref{sec:methodology} explains the methodology adopted in this paper for achieving the objectives previously defined. All experimental configuration and some assumptions needed to reproduce this work are contained in Section~\ref{sec:experiments}. Section~\ref{sec:results} reports the results found using the proposed methods and discusses them. Finally, Section~\ref{sec:conclusion} presents our final remarks and future works.

%% file: sec2_related_works.tex
\section{Related Work}
\label{sec:related_work}

This section presents background knowledge and a literature review on open set and semantic segmentation methods for image classification and semantic segmentation. Most of the open set techniques developed today are adaptations form closed set methods.

\subsection{Open Set Classification}
\label{sec:related_open_set}

In \cite{scheirer2013toward} the authors 
describe an open set version of the well-known Support Vector Machine (SVM) algorithm developed to classify scenes. The choice of using SVM was made because it has various alluring characteristics that can help in this scenario: its answers are global and unique; it has a basic geometric understanding, and it does not rely upon the dimensionality of the information space. 
\cite{bendale2015towards} present and develop a technique for an ``open world'', a recognition system that should update new object categories and be robust to these unseen groups, in addition, to have minimum downtime. To do so, its first step is to continuously detect novel classes; The second is to update the system to include these new classes when novel inputs are found. \cite{bendale2015towards} propose the Nearest Non-Outlier (NNO) algorithm that evolves a model efficiently by adding object categories incrementally while detecting outliers and managing open space risk.

Looking for a deep learning solution, \cite{bendale2016towards} show a new model, called OpenMax, that represents an alternative for the SoftMax function as the final layer of the network, which estimates the probability of an input being from an unknown class. Reducing the number of errors made by a deep network when given fooling generated images. Using a shallow approach, \cite{junior2017nearest} propose a method named Open-Set NN (OSNN) and a variation called OSNNcv, both are able to recognize samples from unknown classes during training time and outperform other approaches in the literature. OSNNcv method verifies if the test sample can be classified as unknown, checking if the two closest samples are from different classes. The OSNN method uses the ratio of similarity scores to the two most similar classes by applying a threshold on it. One of the advantages of this approach is that it is inherently multiclass, which means it is not affected as the number of classes for training increases. OSNN has the characteristic of being inherently multi-class (non-binary-based), differently from other state-of-the-art approaches. Usually these last ones lose some efficiency when the number of classes is increased, while the method proposed by \cite{junior2017nearest} is not affected by the number of classes. 

\subsection{Semantic Segmentation}
\label{sec:related_segmentation}

Besides its open set characteristics, our method relies heavily on deep convolutional architectures. 
Convolutional Neural Networks (CNNs) \cite{krizhevsky2012imagenet,simonyan2014very,szegedy2015going,he2016deep,huang2017densely} have been the state-of-the-art method for most Computer Vision classification tasks for the better part of the last decade. The first widely adopted CNN was AlexNet \cite{krizhevsky2012imagenet}, followed by the deeper VGG \cite{simonyan2014very} and GoogLeNet \cite{szegedy2015going} architectures. 

Between 2015 and 2017, it was observed that the training of earlier layers was severely hampered in deeper architectures due to the Vanishing Gradient problem. Residual Networks (ResNets) \cite{he2016deep} and Densely Connected Convolutional Networks (DenseNets) \cite{huang2017densely} were designed to deal with this limitation of simply stacking convolutions on top of each other by employing shortcuts for the backward gradient between shallower and deeper layers. 
CNNs can be used for image segmentation by classifying the central pixel of a region according to its class and iterating this algorithm across all pixel positions, as will be further explained in Section~\ref{sec:methodology}.

\subsection{Gaps Explored by the Proposed Methods}
\label{sec:related_gaps}


The knowledge gap explored by this paper becomes clear when observing the related works presented in this section. None of the existing methods properly perform both open set inference and semantic segmentation. Therefore, as far as the authors are aware, the methods proposed in this work are the first techniques that can semantically segment a remote sensing imagery on an open set scenario.


%% file: sec3_methodology.tex
\section{Methodology}
\label{sec:methodology}


This section describes in detail the two proposed methods: (1) OpenPixel, presented in Section~\ref{sec:pixelwise}; and (2) Morph-OpenPixel, presented in Section~\ref{sec:morph_openpixel}.
Both methods are based on existing closed set methods for semantic segmentation \cite{nogueira2016learning}. 

\subsection{OpenPixel: Pixelwise Open Set Classification}
\label{sec:pixelwise}

\begin{figure}[h!]
    \includegraphics[width=\columnwidth]{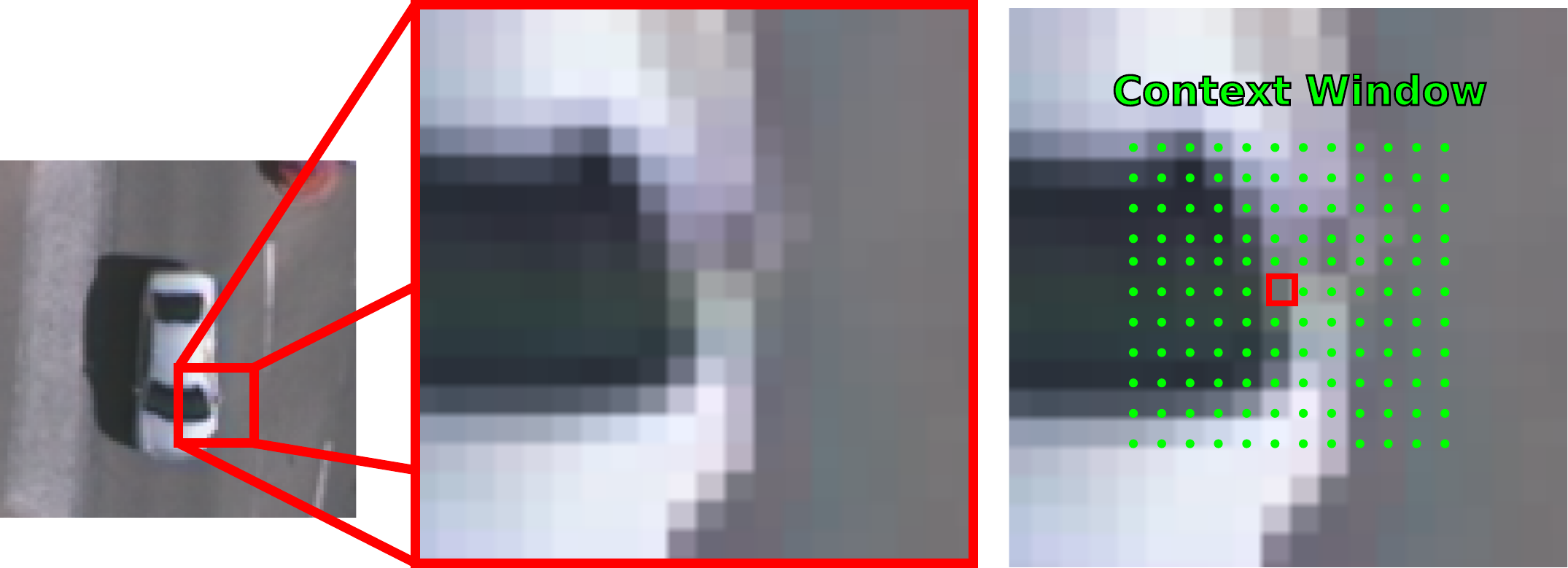}
    \caption{Context window used to evaluate each pixel in the Pixelwise network. All the pixels marked in green are used as the context of the pixel marked as red.}
    \label{fig:context-window}
\end{figure}

The first method proposed in this paper is an open set adaptation from the closed set Pixelwise algorithm proposed by \cite{nogueira2016learning}. This Pixelwise approach consists of the individual treatment of all the pixels present in the images.
Since the pixel itself has not enough information to allow its classification, context windows are employed.
Precisely, context windows are crops of $55 \times 55$ pixels with the central one representing the crop class.
By iterating over all pixels in an image, it is possible to train a CNN
for patch-level classification by using the contextual window.
Figure~\ref{fig:context-window} depicts the context window and central pixel used in this patch-wise classification process.

\begin{figure*}[!ht]
    \centering
    \renewcommand{\currprop}{\textwidth}
    \includegraphics[width=\currprop]{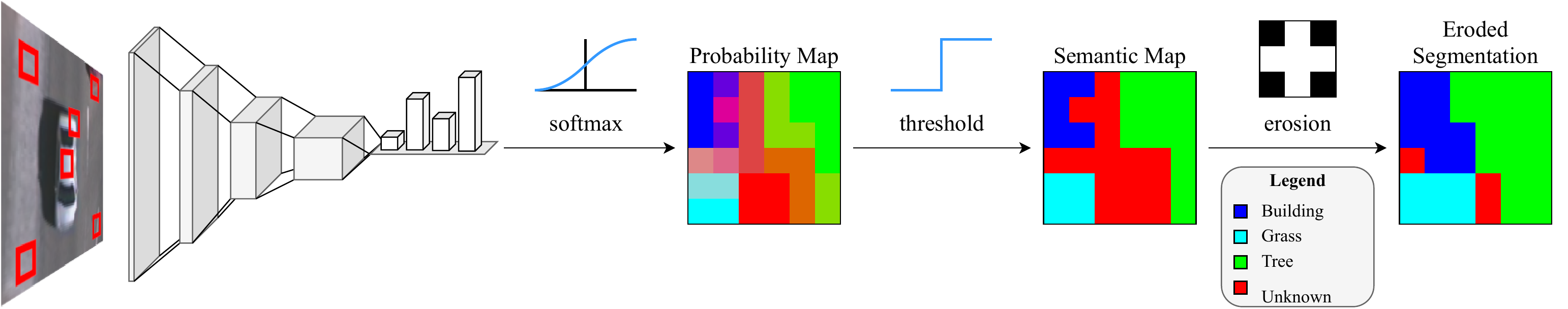}
    \caption{Simplified OpenPixel and Morph-OpenPixel architecture. The distinction between the networks is that OpenPixel does not contain the last morphological filtering step.}
    \label{fig:arch_open}
\end{figure*}

In order to compose OpenPixel, we added an extra layer at the end of the closed set CNN responsible for thresholding each pixel-wise prediction as well as a layer to filter False Positives, as shown in Figure~\ref{fig:arch_open}.
This network receives an image as input and processes it using three layers, each one composed of convolutional operation, Rectified Linear Unit (ReLU) activation, and max pooling. Then, three Fully Connected (FC) layers further process the activation maps from the convolutional blocks to classify each patch's central pixel as pertaining to some class.
Details of this architecture can be seen in Table~\ref{tab:openpixel_arch}. 

\begin{table}[!ht]
    \centering
    \caption{OpenPixel architecture layers and parameters.}
    \resizebox{\columnwidth}{!}{
        \begin{tabular}{|c|c|c|c|}
        \hline
        \textbf{Layer}      & \textbf{Activation Channels} & \textbf{Field of View} & \textbf{Stride} \\ \hline
        \textbf{2D Conv}    & $3 \rightarrow 64$           & $4 \times 4$           & 2               \\ \hline
        \textbf{2D Pooling} & $64 \rightarrow 64$          & $2 \times 2$           & 2               \\ \hline
        \textbf{2D Conv}    & $64 \rightarrow 128$         & $4 \times 4$           & 1               \\ \hline
        \textbf{2D Pooling} & $128 \rightarrow 128$        & $2 \times 2$           & 2               \\ \hline
        \textbf{2D Conv}    & $128 \rightarrow 256$        & $2 \times 2$           & 2               \\ \hline
        \textbf{2D Pooling} & $256 \rightarrow 256$        & $2 \times 2$           & 1               \\ \hhline{|=|=|=|=|} 
        \textbf{Layer}      & \multicolumn{3}{c|}{\textbf{Input/Output Dimensions}}                   \\ \hline
        \textbf{FC}         & \multicolumn{3}{c|}{$256 \rightarrow 1024$}                             \\ \hline
        \textbf{FC}         & \multicolumn{3}{c|}{$1024 \rightarrow 1024$}                            \\ \hline
        \textbf{FC}         & \multicolumn{3}{c|}{$1024 \rightarrow 1024$}                            \\ \hline
        \textbf{FC}         & \multicolumn{3}{c|}{$1024 \rightarrow N_{classes}$}                     \\ \hline
        \end{tabular}
    }
    \label{tab:openpixel_arch}
\end{table}





This architecture is the same as Pixelwise for the closed set scenario, adding a probability threshold after the softmax. To do so, a pixel with a class confidence (given by the softmax) that exceeds a determined threshold is labeled as belonging to that class.
However, if the pixelwise probability is inferior to the threshold, the pixel is classified as unknown. As the value of probability given by the softmax varies between 0 and 1, the possible values of threshold also vary between 0 and 1.

\subsection{Morph-OpenPixel: Morphological Filtering}
\label{sec:morph_openpixel}

After the result predicted at the softmax layer and applied the threshold by the network, a post-processing morphological filter is applied at the pixels classified as unknown.
This filter analyses the neighbors of each unknown pixel to determine if it belongs to a border or if it is an inside pixel. If it belongs to the border, it has neighbors from other classes. In this case, the classification is exchanged to the label with a higher amount of pixels in the neighborhood.
If all the pixels are from the unknown class, it means the central pixels are not on the border and it should remain labeled as unknown.

The applied filter can be seen as an erosion done over the unknown pixels, with the only difference being its adaptation for a multi-class context.
According to \cite{gonzalez2000processamento}, the process of erosion of a set $A$ by another set $B$ (both in $Z^{2}$) is defined as the set of all points in $z$ such that $B$, translated by $z$, is contained in $A$, as represented on the Equation~\ref{eq:erosion}:
\begin{equation}
    A \ominus B = \left \{ z | (B)_{z} \subseteq A\left.  \right \}\right.
\label{eq:erosion}
\end{equation}

Since this process of erosion is only applied over the pixels classified as unknown by the network, the technique reduces the amount of false unknown labels created by the uncertainty of boundary regions.
The rightmost modules in Figure~\ref{fig:arch_open} shows a didactic example of the application of the morphological filter.



%% file: sec4_experiments.tex
\section{Experimental Setup}
\label{sec:experiments}

In this section, we introduce the configuration used during the experiments and needed to guarantee the reproducibility of results. Section~\ref{sec:dataset} describes the Vaihingen dataset used in our experimental procedure, Section~\ref{sec:protocol} presents the protocol for training and testing and, finally, Section~\ref{sec:metrics} introduces all the metrics used for quantitative evaluation.

\subsection{Dataset}
\label{sec:dataset}


The Vaihingen dataset\footnotemark \footnotetext{\url{http://www2.isprs.org/commissions/comm3/wg4/2d-sem-label-vaihingen.html}} contains 33 patches of different sizes, each consisting of a True OrthoPhoto (TOP) extracted from a larger TOP mosaic.
These 33 patches were captured over the city of Vaihingen in Germany by the German Society for Photogrammetry and have a ground sampling distance of 9 cm. The Ground Truth consists of 5 classes: street, building, grass, tree, and car. 


The dataset was created to be well controlled and avoid areas without data. To do so, the patches were selected from the central part of the mosaic and not from the boundaries. Even with this approach, some small missing information could occur, to prevent that to happen, interpolation is used to fill all the gaps. The TOP is 8 bit TIFF files with three bands, being three RGB bands, corresponding to the near-infrared, red and green. 

\subsection{Training/Predicting Protocol}
\label{sec:protocol}

For the Vaihingen dataset, we followed the protocol proposed by \cite{nogueira2019dynamic}.
Precisely, the dataset is divided into two sets, one for training and one for testing. The testing set consists of images from the patches 11, 15, 28, 30 and 34, whereas the remaining images are used as training. 

\begin{figure}[ht!]
    \centering
    \renewcommand{\currprop}{0.3\columnwidth}
    \subfloat[Image]{
        \includegraphics[width=\currprop]{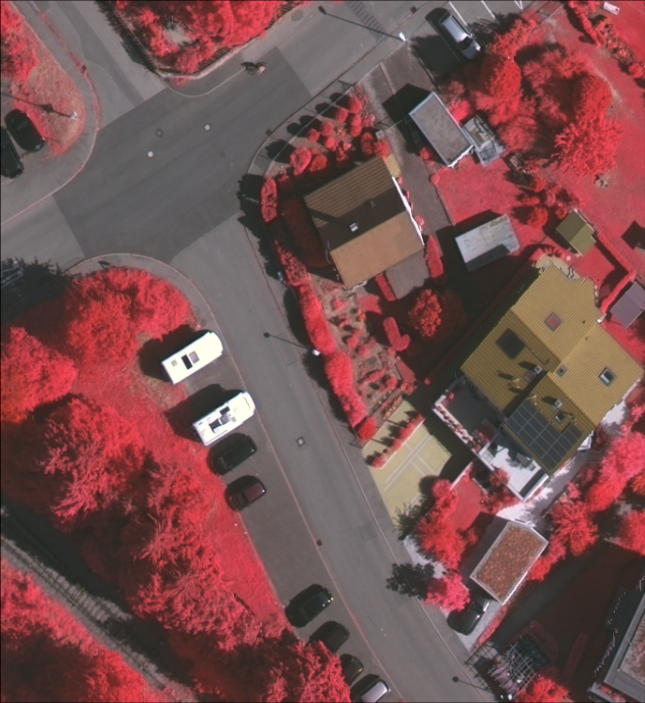}%
        \label{fig:protocol-unknown-original}
    }
    \subfloat[Unknown Building]{
        \includegraphics[width=\currprop]{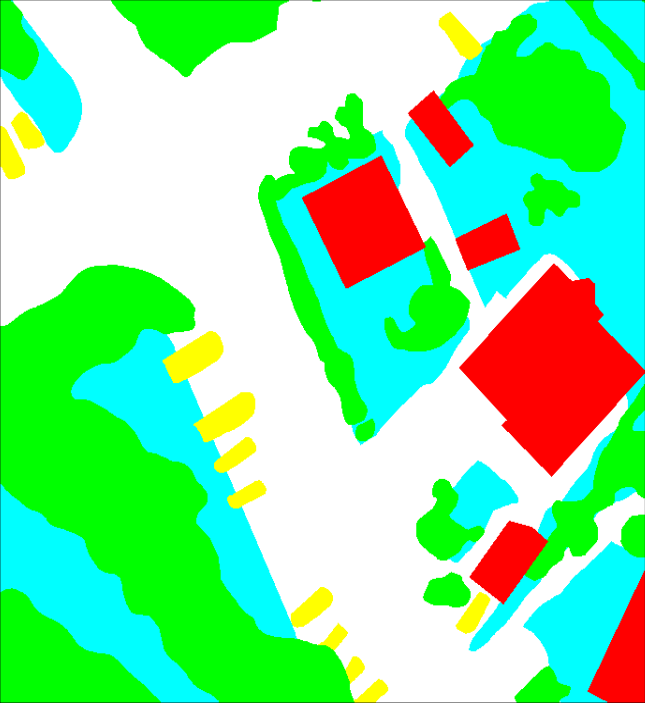}%
        \label{fig:protocol-unknown-building}
    }
    \subfloat[Unknown Car]{
        \includegraphics[width=\currprop]{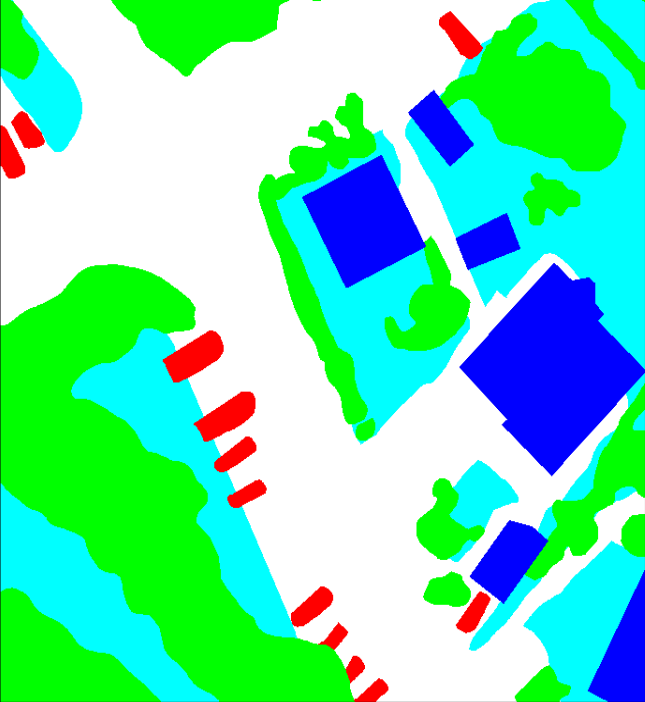}%
        \label{fig:protocol-unknown-car}
    }
	\hfil
    \subfloat[Unknown Grass]{
        \includegraphics[width=\currprop]{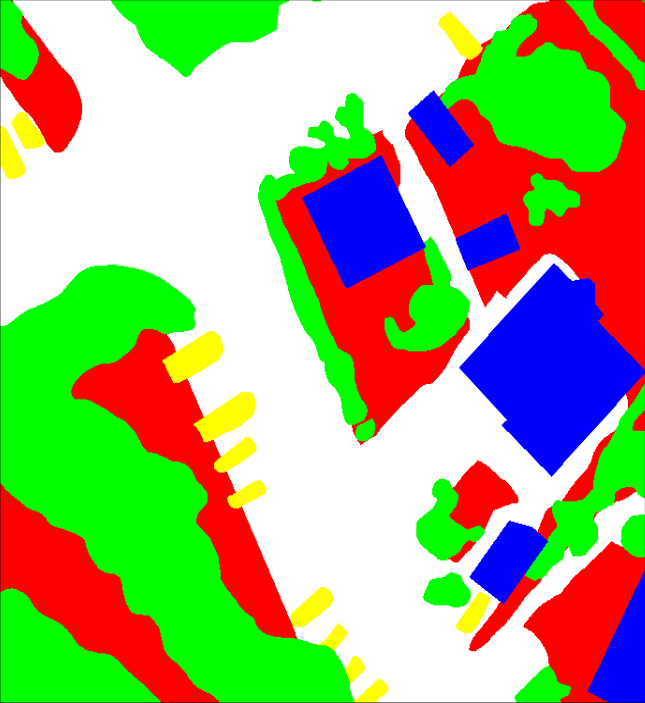}%
        \label{fig:protocol-unknown-grass}
    }
    \subfloat[Unknown Street]{
        \includegraphics[width=\currprop]{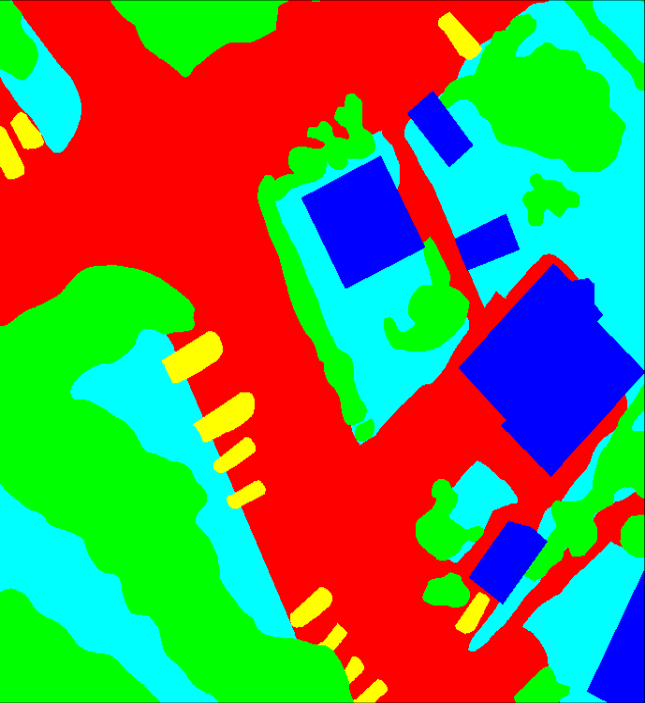}%
        \label{fig:protocol-unknown-street}
    }
    \subfloat[Unknown Tree]{
        \includegraphics[width=\currprop]{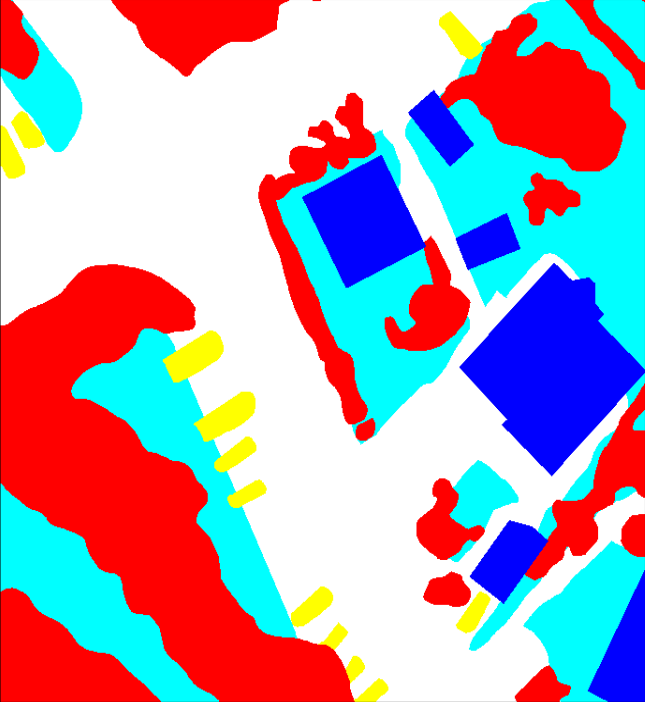}%
        \label{fig:protocol-unknown-tree}
    }
    \caption{Representation of the protocol used to vary the unknown class during the experiments.}
    \label{fig:protocol-unknown}
\end{figure}

During the experiments, 4 out of the 5 labels of the dataset are used as known classes (employed to train the model) while the remaining one is exploited as the unknown class.
To better evaluate the proposed method, the protocol also varied the class used as unknown.
In each iteration of the experiments, a different class of the dataset was not fed to the algorithm during the training stage, only being seeing during the testing.
Figure~\ref{fig:protocol-unknown} represents the variation on the unknown class.


The validation set, only used by the OpenPixel method to determine the best threshold values, is composed of a part of images from the training set.
When this set is used, the method is trained considering a set of images consisted of the training set, excluding examples on the validation set.

In the predicting phase, each input image is processed independently by the trained deep model, which outputs a final prediction map in which for each pixel a label indicates whether or not the pixel belongs to a known class, and if it belongs, to which class. This stage uses and presents the one class not seen during learning as the unknown class.

It was analyzed four different contexts with the OpenPixel network.
The first one was training and testing the networks as the traditional closed set.
In this case, the methods are trained and evaluated considering all existing classes of the dataset.
This context was only used to show the accuracy of the method without the open set concept.

The second context was training the model as a closed set, but testing it in an open set scenario.
In this scenario, the model must classify, during the testing phase, pixels from classes that do not exist training set, resulting in an misclassification of such samples.
This scenario, along the next one, shows the relevance of the open set concept for semantic segmentation.

The third one is training and testing the method in open set scenarios.
In this case, the network knows it will analyze some pixels from not known classes during the training phase and will be able to classify them as unknown. 

The last one is very similar to the third one, the only difference is that it applies the morphological filter to enhance the prediction and mitigate some False Positives on the OpenPixel methodology.

\subsection{Metrics}
\label{sec:metrics}

All results obtained in this work are reported using Cohen’s Kappa Index, Overall and Normalized Accuracy scores, given that these metrics take into account the existence of multiple classes and the importance of correct segmenting all of them \cite{congalton2008assessing}. 
The Kappa index ($\kappa$) measures the agreement between the reference map and the predicted outcome.
This is a common metric employed on dataset with imbalanced classes, such as the remote sensing ones \cite{dos2013semi}.
The accuracy or Overall Accuracy (OA) is a common metric used to infer the correctness of a method in classification tasks. 
One problem that can affect the value of an overall accuracy is the unbalance of a testing example. 
Therefore, it is common to complement this metric by using the Normalized Accuracy (NA), which takes in account the imbalance for each class. The NA can be seen as the combination of the accuracies for each class.


%% file: sec5_results.tex
\section{Results and Discussion}
\label{sec:results}

The conducted experiments (and results) presented in this section
aim to answer the following questions: (1) What are the best configurations of parameters for the OpenPixel method? (2) Is the OpenPixel able to semantic segment a remote sensing image?

\begin{figure*}[!ht]
    \centering
    \renewcommand{\currprop}{0.3\textwidth}
    \subfloat[Pixelwise]{
        \includegraphics[width=\currprop]{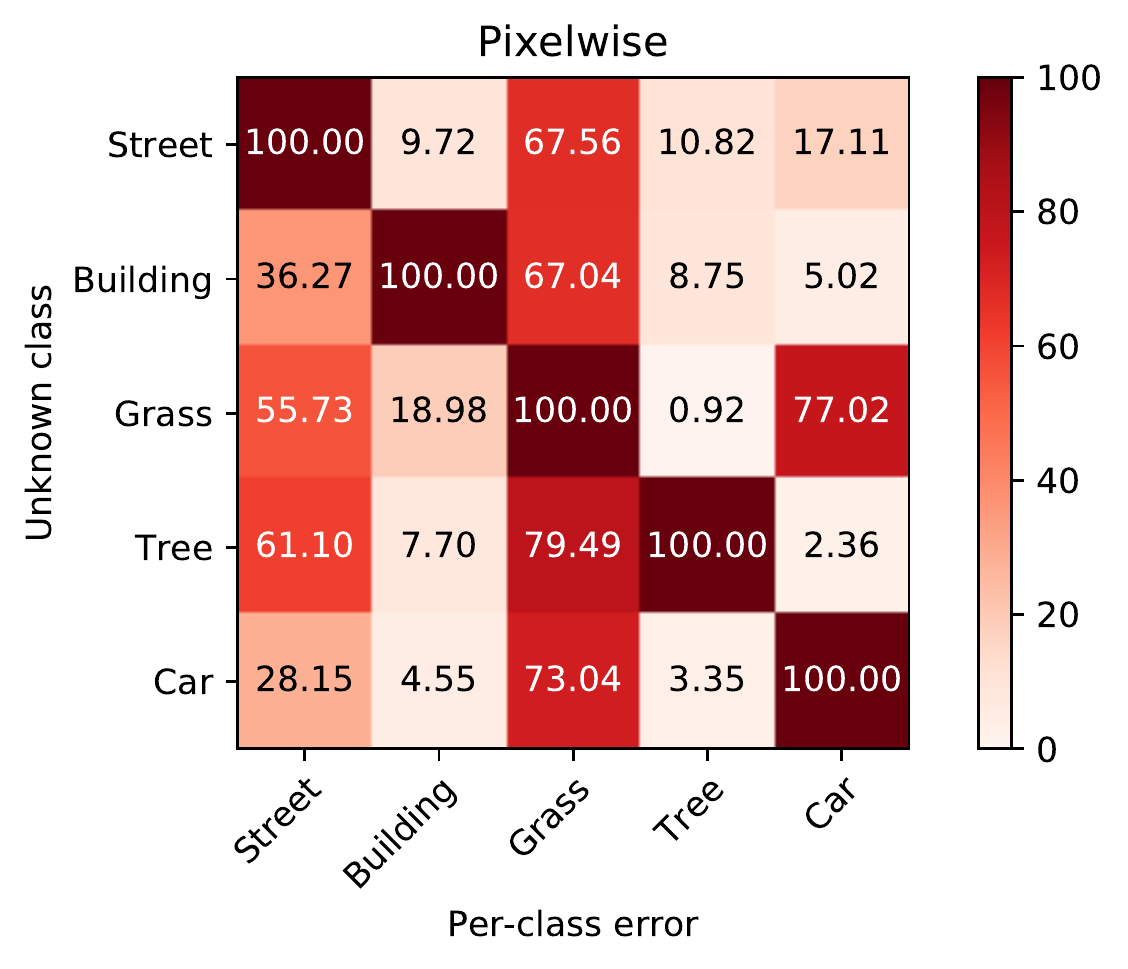}%
        \label{fig:not_cm_1}
    }
    \hfil
    \subfloat[OpenPixel]{
        \includegraphics[width=\currprop]{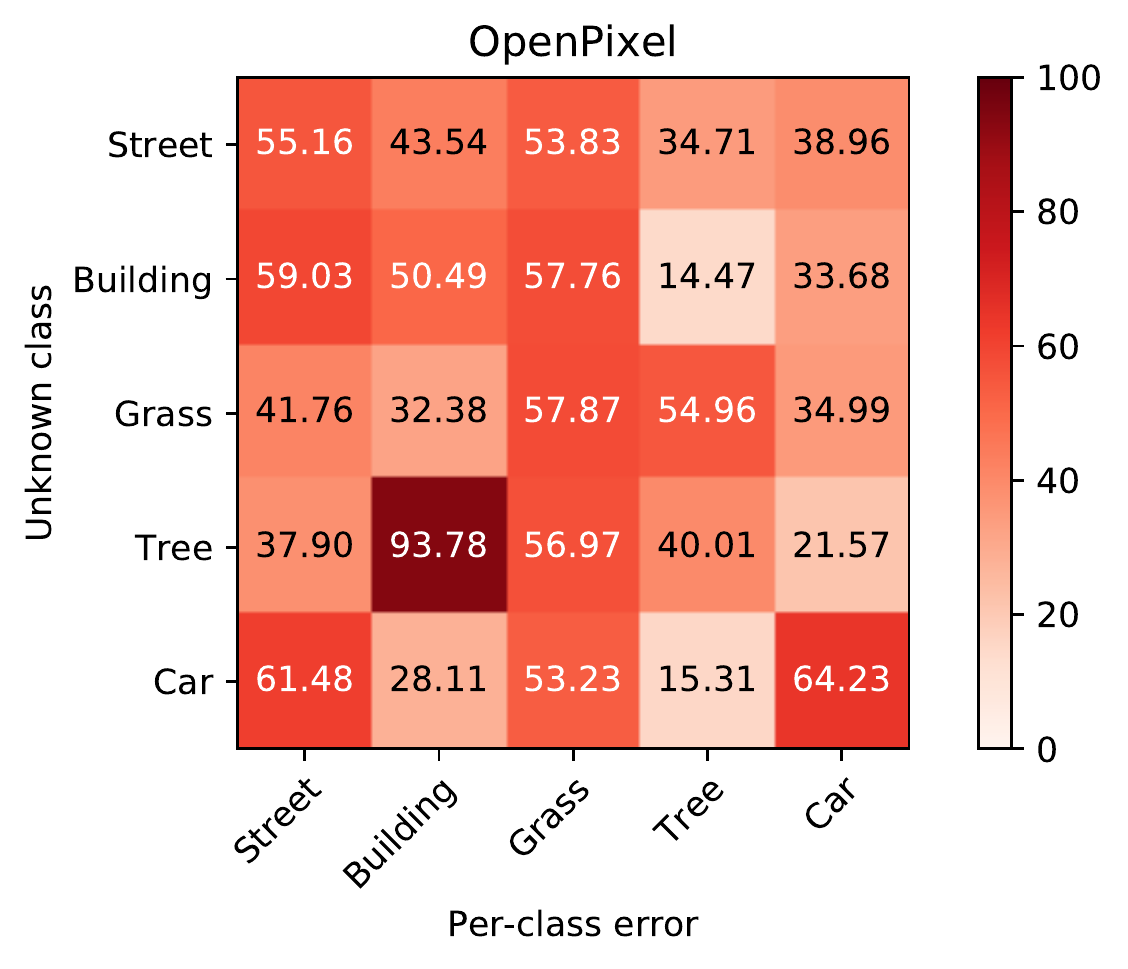}%
        \label{fig:not_cm_2}
    }
    \hfil
    \subfloat[Morph-OpenPixel]{
        \includegraphics[width=\currprop]{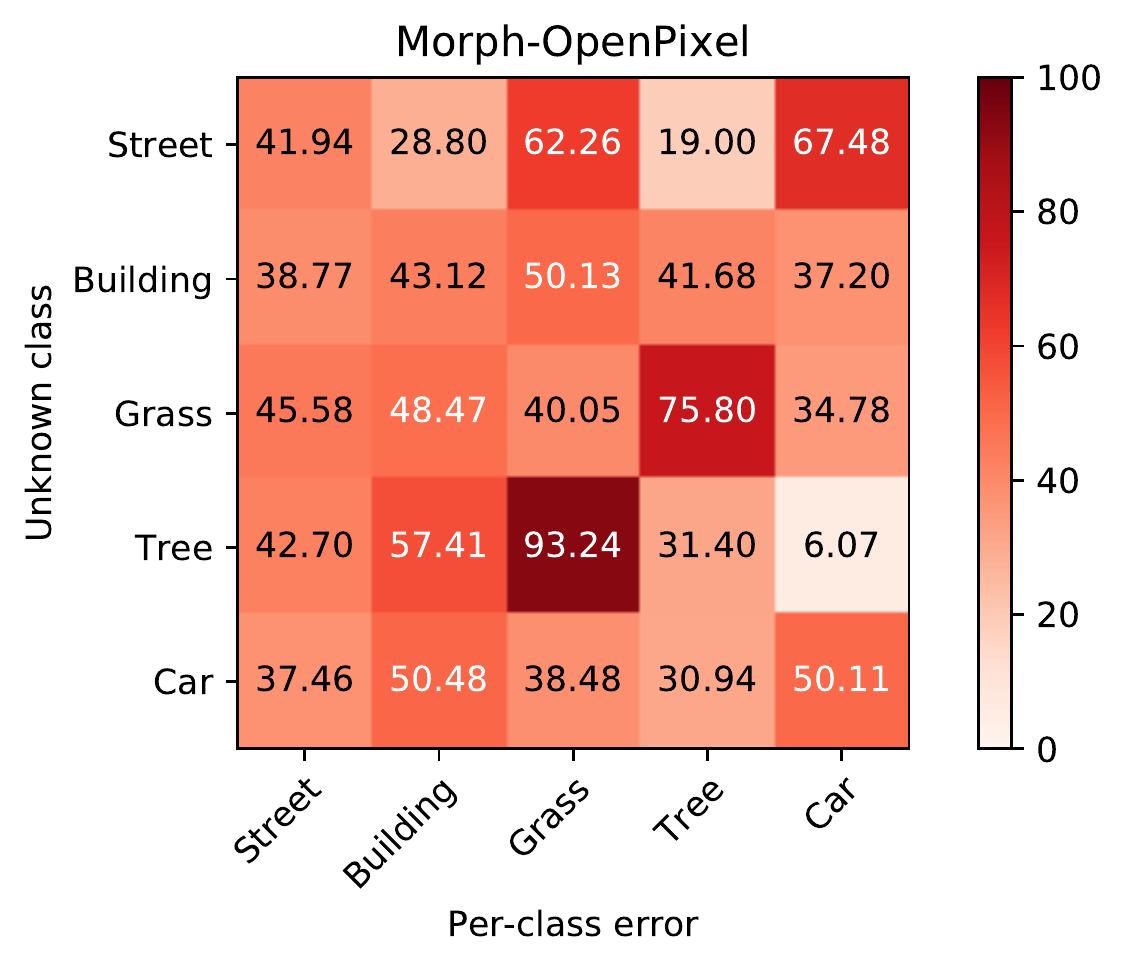}%
        \label{fig:not_cm_3}
    }
    \caption{Per-class error rates for each unknown class according to the closed set baseline (a) and the proposed open set methods (b-c).}
    \label{fig:not_a_confusion_matrix}
\end{figure*}

\begin{figure}[!ht]
    \centering
    \renewcommand{\currprop}{\columnwidth}
    \includegraphics[width=\currprop]{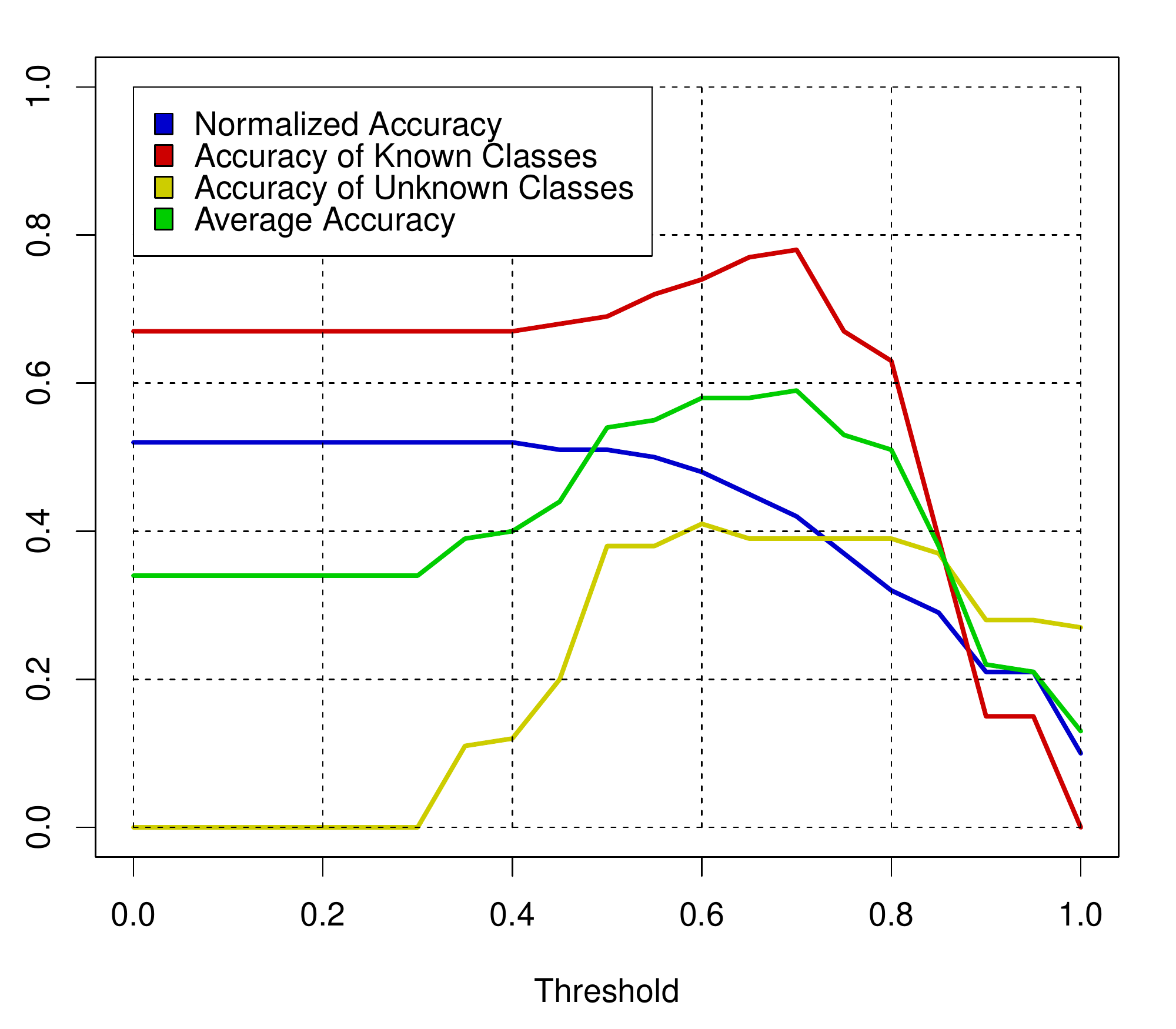}
    \caption{Plot representing the variation in threshold values for OpenPixel and its impact in evaluation metrics.}
    \label{fig:threshold-graph}
\end{figure}

\begin{figure}[ht!]
    \centering
    \renewcommand{\currprop}{0.465\columnwidth}
    \subfloat[RGB Image]{
        \includegraphics[width=\currprop]{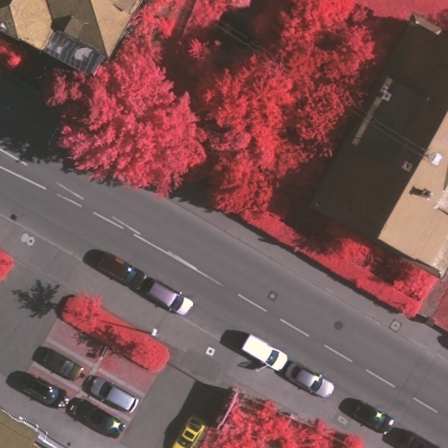}%
        \label{fig:unknown_classes_rgb}
    }
    \subfloat[Ground Truth]{
        \includegraphics[width=\currprop]{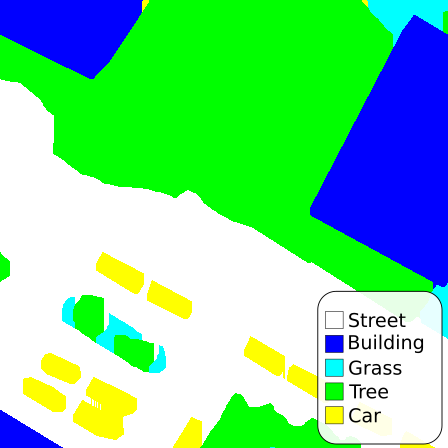}%
        \label{fig:unknown_classes_gt}
    }
    \hfil
    \subfloat[Pixelwise Closed Set Prediction]{
        \includegraphics[width=\currprop]{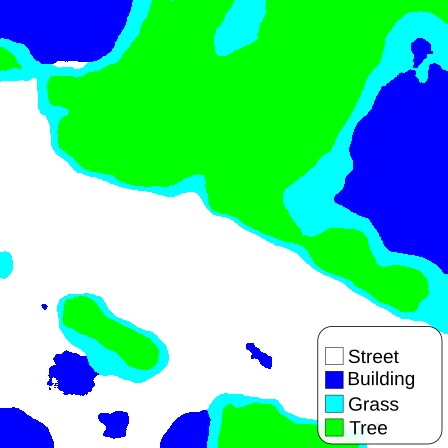}%
        \label{fig:unknown_classes_closed}
    }
    \subfloat[Morph-OpenPixel Prediction]{
        \includegraphics[width=\currprop]{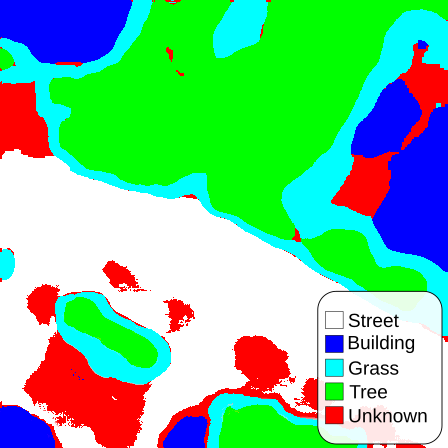}%
        \label{fig:unknown_classes_morph}
    }
    \caption{Example wherein Pixelwise wrongly classifies the unknown class (car) as known, while the Morph-OpenPixel classifies the areas as unknown.}
    \label{fig:unknown_classes}
\end{figure}

\subsection{OpenPixel Configuration}
\label{sec:pixel-config}

In order to define the best configuration for the OpenPixel technique, a set of experiments were conducted analyzing the effect of the threshold in the final result.
In order to evaluate these variations it was considered four metrics: the normalized accuracy over all samples, over instances of known classes, over samples of the unknown class, and an arithmetic median between the accuracy of known and unknown instances.

The normalized accuracy over all samples gives an idea of the results expected when using those values of threshold, but since the number of known classes is higher than the unknown classes, this metric may not be enough for finding the best parametrization. 
The accuracy of instances of known classes is measured only taking into account the classes seen by the method during learning. The unknown accuracy, is the opposite to what was described above. It only uses the pixels belonging to the class not learned by the algorithm.
Finally, the arithmetic median (between the accuracy of known and unknown instances) has the goal to find the best balance between them.
Figure~\ref{fig:threshold-graph} presents the graph showing the different values of the accuracies described here when altering the threshold used by the OpenPixel method.

Observing mainly the values of the Average Accuracy presented in Figure~\ref{fig:threshold-graph}, the optimal threshold was empirically found to be 0.7. This value was used for both OpenPixel and Morph-OpenPixel in all further tests.
One should notice that this optimal threshold was computed globally according to the performance of all classes, even though it is possible to tune this value individually by class. 
However, we chose not to make a per-class evaluation due to high computational time, given that, in this case, for each class, a network would have to be trained from scratch.
Note, however, that although expensive to train, this simple threshold idea combined with the pixelwise network produces a method that takes less to obtain results when compared to other open set techniques.
Specifically, the OpenPixel approach can be trained in 8 hours (according to the setup configuration used in this paper) and is able to generate a prediction for all the images on the testing set in under than 30 minutes.

\subsection{OpenPixel Evaluation}
\label{sec:results_openpixel}

Considering the analysis performed on the previous section, we analyze the effectiveness of the proposed method to perform open set semantic segmentation.
Table~\ref{tb:pixelwise} presents the obtained results.
Through the table, it is possible to note that the proposed method, OpenPixel, yielded acceptable results, showing the feasibility of combining semantic segmentation and open set concepts. 

\begin{table}[h!]
    \centering
    \caption{Normalized Accuracy and Kappa Index obtained by the OpenPixel method and baselines.}
    \tabcolsep=0.155cm
    \resizebox{\columnwidth}{!}{

        \begin{tabular}{cccc}
        \hline
        \textbf{Network}                   & \textbf{\begin{tabular}[c]{@{}c@{}}Overall\\ Accuracy\end{tabular}} & \textbf{\begin{tabular}[c]{@{}c@{}}Normalized\\ Accuracy\end{tabular}} & \textbf{Kappa}             \\ \hline
        Pixelwise (closed)                          & 55.84\%                                                             & 53.98\%                                                                & 0.5585                     \\
        OpenPixel                           & 55.78\%                                                             & 53.15\%                                                                & 0.5106                     \\
        Morph-OpenPixel &  \textbf{57.51\%}                                         & \textbf{54.23}\%                                            & \textbf{0.5602} \\ \hline
        \end{tabular}
}
    \label{tb:pixelwise}
\end{table}

Analyzing the results shown in Table~\ref{tb:pixelwise}, one can observe that the OpenPixel method trained on an open set configuration achieved similar results to the Pixelwise closed set when tested on an open set scenario.
Even though the open set methods only surpassed the closed set architecture with morphology filtering (Morph-OpenPixel) the simpler version (OpenPixel) also has the benefit of finding unknown classes that are always mislabeled by closed set methods in this scenario. 

This advantage is better demonstrated in Figure \ref{fig:not_a_confusion_matrix}, in which the Error Rate is reported.
Observing the error rate of the Pixelwise network, more precisely the main diagonal, one can notice that in all cases the error is 100\%, an expected outcome, since the network was not designed to deal with unknown classes, misclassifying all the pixels belonging to this class.
When comparing the main diagonal from all three methods, it is possible to understand the advantage of using an open set semantic segmentation technique.
Both proposed methods have error rate lower than the Pixelwise network for the unknown classes, while keeping the error rate low on most known classes.

\begin{figure}[h!]
    \centering
    \renewcommand{\currprop}{\columnwidth}
    \includegraphics[width=\currprop]{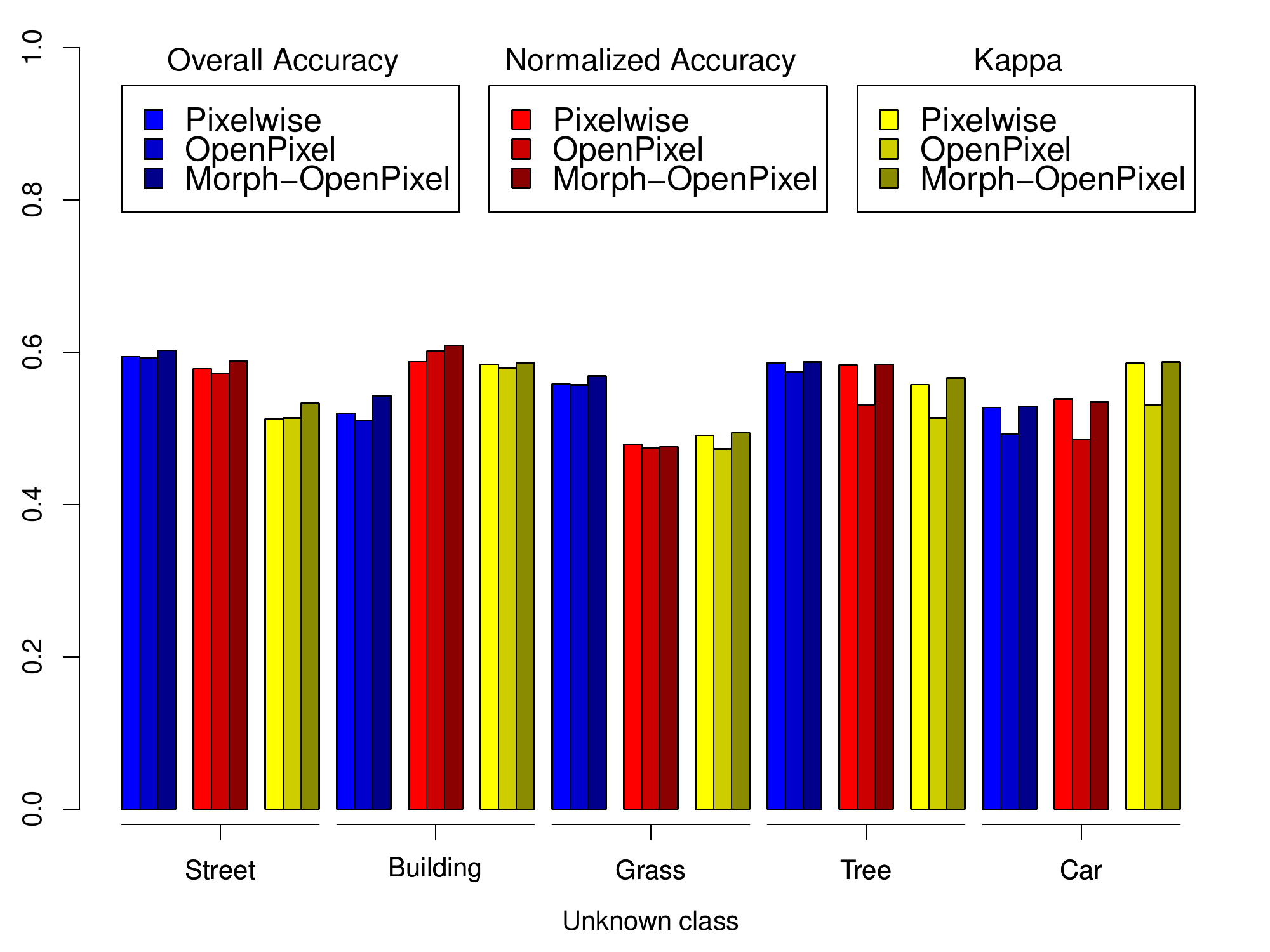}
    \caption{Overall Accuracy, Normalized Accuracy and Kappa Index obtained by the Pixelwise network and the proposed methods, OpenPixel and Morph-OpenPixel,  for each unknown class in the experimental procedure.}
    \label{fig:pixel-accuracy-graph}
\end{figure}

Aside from this, since each class has distinct patterns,
it is essential to evaluate the proposed method by varying the unknown class.
Results for this experiment are presented in Figure~\ref{fig:pixel-accuracy-graph}.
Through the table, it is possible to note that Morph-OpenPixel achieved better metric rates on most of the setups, including the ones in which all the networks had a worse performance.
The gap between the results achieved by the Morph-OpenPixel and the OpenPixel is another evidence to understand the importance of applying a morphological filter to mitigate the uncertainty existing on boundary regions. 

Visual results are presented in Figure~\ref{fig:unknown_classes}, in which the ground truth had known and unknown classes, being the car class (in yellow) the unknown. The Closed Set Pixelwise technique wrongly classified all the pixels belonging to cars, as was expected, since it does not know this class. The prediction resulting from the Morph-OpenPixel, as it can be noticed, has most of the instances of the known classes classified correctly, while still classifying the car pixels as unknown (in red).

Moreover, it is possible to note that the main reason for the low accuracy is the existence of shadow in the images. While images from this dataset present those shadows, the ground truths do not make any distinction, labeling the same as an area of the class without shadow, and in the prediction phase, the open set method classifies the darker areas as unknown classes, a different result as the one expected.

%% file: sec6_conclusion.tex
\section{Conclusion and Future Works}
\label{sec:conclusion}

Open set scenarios are more robust for modeling the real world, since a fully controlled case where all the possible classes are known in advance is hardly going to be found in practice.
This is even more prevalent in remote sensing images, that can present a high variability of classes, as distinct types of vegetation and cars or even people.
For this reason is important to develop open set semantic segmentation methods, i.e., approaches capable of correctly classifying pixels from known classes while identifying samples from unknown labels.

One of the proposed methods (OpenPixel) produced acceptable rates of normalized accuracy when compared to closed set methods on the same dataset.
On average, OpenPixel achieved an overall accuracy of 57.51\%, a normalized accuracy of 54.23\%, and a Kappa Index of 0.4600.
Observing the experiments and the results presented in this paper, it is possible to affirm that the proposed methods are promising in the task of semantic segmenting pixels belonging to unknown classes, while still correctly classifying most of the pixels from known classes.


In conclusion, this paper main contributions are: (1) an initial investigation resulting in the first evidences that indicates the feasibility of combining semantic segmentation and open set concepts;
and (2) the development of two methods for open set semantic segmentation. 

As future works, the Pixelwise method presented can be incremented by adding more techniques of data augmentation, as histogram equalization, or brightness control can be used to mitigate the shadow problem encountered.